\documentclass[10pt,twocolumn,letterpaper]{article}

\usepackage{cvpr}
\usepackage{times}
\usepackage{epsfig}
\usepackage{graphicx}
\usepackage{amsmath}
\usepackage{amssymb}
\usepackage{mathtools}
\usepackage{url}
\usepackage[backend=bibtex]{biblatex}
\usepackage{mathtools}
\DeclarePairedDelimiter\floor{\lfloor}{\rfloor}
\usepackage{tabularx}
\usepackage{algorithm}
\usepackage[noend]{algpseudocode}
\makeatletter
\newcommand{\multiline}[1]{%
  \begin{tabularx}{\dimexpr\linewidth-\ALG@thistlm}[t]{@{}X@{}}
    #1
  \end{tabularx}
}
\makeatother
\cvprfinalcopy % *** Uncomment this line for the final submission

\def\cvprPaperID{8682} % *** Enter the CVPR Paper ID here

\begin{document}

%%%%%%%%% TITLE
\title{A Method for Arbitrary Instance Style Transfer}

\author{Zhifeng Yu\thanks{Primary Author}\\
Lionsgate Entertainment Corp.\\
{\tt\small yuzhifeng1111@gmail.com}
\\
\and
Yusheng Wu\thanks{Primary Author}\\
University of Southern California\\
{\tt\small yushengw@usc.edu}
\\
\and 
Tianyou Wang\thanks{We enjoyed and appreciated the discussions with Tianyou and are thankful to the contributions he had in the paper writing process}\\
University of Southern California\\
{\tt\small tianyouw@usc.edu}
}

\title{A Method for Arbitrary Instance Style Transfer}
\maketitle
%\thispagestyle{empty}

%%%%%%%%% ABSTRACT
\begin{abstract}
  The ability to synthesize style and content of different images to form a visually coherent image holds great promise in various applications such as stylistic painting, design prototyping, image editing, and augmented reality. However, the majority of works in image style transfer have focused on transferring the style of an image to the entirety of another image, and only a very small number of works have experimented on methods to transfer style to an instance of another image. Researchers have proposed methods to circumvent the difficulty of transferring style to an instance in an arbitrary shape. In this paper, we propose a topologically inspired algorithm called Forward Stretching to tackle this problem by transforming an instance into a tensor representation, which allows us to transfer style to this instance itself directly. Forward Stretching maps pixels to specific positions and interpolate values between pixels to transform an instance to a tensor. This algorithm allows us to introduce a method to transfer arbitrary style to an instance in an arbitrary shape. We showcase the results of our method in this paper.\footnote[1]{Our results are color images. Please adjust printer or PDF Viewer for best reading experience} Code will be made available in our supplementary material.
\end{abstract}

%\includegraphics[width=3cm, height=4cm]{seated-nude}

%%%%%%%%% BODY TEXT
\section{Introduction}

In recent years, researchers have made substantial progress on applying deep neural networks in the field of image style transfer in which the style of an image, referred to as style image, is transferred to another image, referred to as content image, without distorting the content image’s 
\begin{center}
\begin{figure}[h]
\begin{center}
\includegraphics[width=6cm, height=6.5cm]{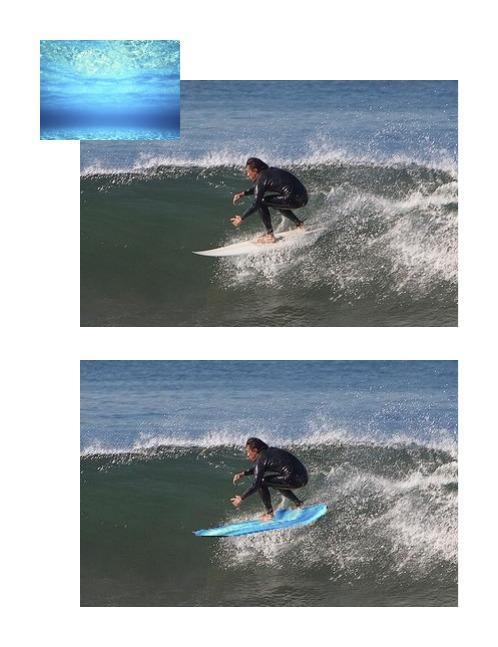}
\caption{\textsf{Instance Style Transfer on Surfboard}}
\end{center}
\end{figure}
\end{center}

\noindent structure. Since the pioneering work of Gatys \etal \cite{Image} in 2016, the goal of most algorithms for image style transfer is to stylize the entirety of the content image. The visually appealing outputs of these algorithms also found their way into the real world. Prisma, one of the most prominent smartphone applications for image editing, was the App of the Year in 2016 on the App Store because of its popularity \cite{Senn}. However, style transfer to the whole content image doesn't realize the full potential of this field in industrial applications. For instance, a fashion designer may want to stylize the shirt within the image for prototyping; a photographer may want to colorize the background of an image for editing. Instance style transfer provides more flexibility to users and can well be extended to other areas such as augmented reality, game development, and more. 

That deep neural network is a collection of matrix operations is a challenge for researchers to develop algorithms for direct instance style transfer because irregular-shaped instances have to be transformed to matrix or tensor

\begin{figure*} 
\centering 
\includegraphics[width=18cm, height = 8.5cm]{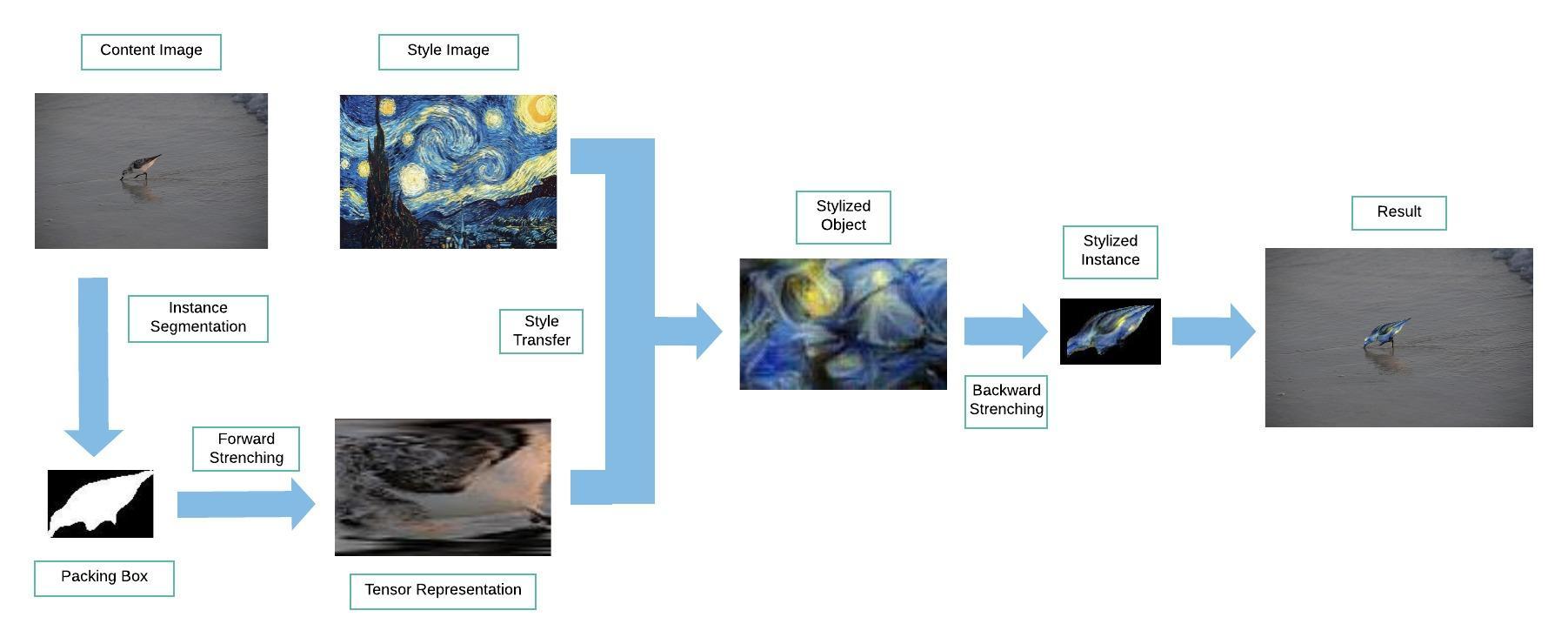} 
\caption{\textsf{The workflow of our method. Please note that Stylized Object and Stylized Instance are not on scale.}} 
\label{fig:ds} 
\end{figure*}

\noindent representations before most of the image style transfer algorithms can be applied. To tackle this challenge, researchers have developed a workaround. Castillo \etal \cite{son} and Kurzman \etal \cite{classbased} stylize the entire content image before extracting the stylized instance and putting this instance back to the original, unstylized content image. 

In this paper, we propose a method that can transfer arbitrary style to an instance in an arbitrary shape. This method comprises the following steps:

1. Segmenting and extracting the instance from the content image 

2. Transforming this instance into tensor representation with our proposed algorithm, Forward Stretching

3. Transferring style to this object

4. Restoring the stylized object into its original shape with Backward Stretching and integrating it back to the content image\\

A visual representation of the method is displayed in Figure.2. 
\section{Related work}
  Since instance segmentation and image style transfer are the backbone of our method, we organize this session into three sub-sessions: Instance segmentation, Image style transfer, and Instance style transfer.
\subsection{Instance segmentation}
  Instance segmentation extracts the instance from the image before we perform instance style transfer. In general, there are two different approaches: proposal-based and segmentation-based. \cite{Mask, Reversible, Instance, Path} are examples of proposal-based approaches where instance segmentation is performed within proposed regions such as feature maps or bounding box proposals. R2-IOS \cite{Reversible} has two sub-networks to recursively and alternately refine the inter-dependent tasks of mask predictions and object proposals in iterations so that improvement of one sub-network will benefit the other. Multitask Network Cascades \cite{Instance} has three sub-networks where downstream tasks (\eg mask prediction) are dependent on upstream tasks (bounding box proposals). These early models are fairly large. In addition, mask predictions and classification are not performed in parallel. Later works, such as Mask R-CNN \cite{Mask} and PANet \cite{Path}, separate mask predictions and classification and use better pooling layer to improve performance and speed. Segmentation-based approaches utilize various techniques to separate individual instances among pixels with the same class label \cite{Deep, SGN}. \cite{Deep} applies Watershed Transform to produce an energy map for pixels in the same semantic class on which instance segmentation is performed. SGN \cite{SGN} predicts breakpoints on pixels in the same semantic class, which are used as boundaries for instance segmentation. 
\subsection{Image style transfer}
  Gatys \etal \cite{Image} approached image style transfer in his influential work as an image reconstruction problem where the stylized image is obtained by iteratively minimizing the distance between a randomly generated image and the content and style representations extracted from a VGG. Such process is fairly slow, and, to address this issue, \cite{perceptual, Texture} introduced neural networks to encode style at training time so that image style transfer can be applied as a feed-forward process at inference time. However, these models have to be retrained for each new style.

  Chen \etal \cite{StyleBank} and Dumoulin \etal \cite{ALearned} proposed models capable of encoding multiple styles. Each style is a group of convolutional filters in \cite{StyleBank} whereas in \cite{ALearned} a particular style is embedded by shifting and scaling activations in the Instance Normalization layer. Still, these models are limited to the styles they have learned. Subsequently, Huang and Belongie \cite{2017} developed a model to transfer arbitrary style to the content image by matching the summary statistics of the content and style features in the Adaptive Instance Normalization (AdaIn) layer. Li \etal \cite{Universal} presented a model for arbitrary style transfer by matching the covariance matrices of feature maps of content and style images in a series of operations called whitening and coloring. 
\subsection{Instance style transfer }
  Both Castillo \etal \cite{son} and Kurzman \etal \cite{classbased} proposed methods using different models to stylize the whole content image first before extracting the stylized instance from the content image and replacing the unstylized instance with the stylized instance in the original content image. Since Castillo \etal \cite{son} used \cite{Image} for image style transfer and Markov Random Field to smooth out the blending of the stylized instance and the unstylized background, this method is fairly slow for industrial applications. Kurzman \etal \cite{classbased} introduced a method that uses \cite{perceptual} for image style transfer, increasing the speed to 16FPS on Tesla P100 GPUs. However, \cite{perceptual} is only capable of encoding one style.
\section{Method}
  In this session, we describe our method in details. To the best of our knowledge, our method is the first one to tackle arbitrary instance style transfer and apply insights from topology to the field of image style transfer, which, we hope, will facilitate future research. 
\subsection{Instance segmentation}
  Our method uses Mask R-CNN \cite{Mask} to extract instance from the content image. Specifically, the backbone of our Mask R-CNN is ResNet-FPN-101 where ResNet \cite{DeepResidual} is the widely used deep neural network for image recognition, FPN \cite{July} is an in-network architecture called Feature Pyramid Network that can be incorporated into ResNet, and 101 is the depth of the ResNet. Such backbone will extract and combine multi-scales features from the input image \cite{July}, providing fine-grained feature representations for mask predictions. After this step, the feature representations will go through a layer called ROI Align to obtain feature maps of same size on which the masks will be predicted \cite{Mask}. These feature maps will be fed into the mask head (and other heads, but we are only concerned about the masks), which is a collection of convolutional filters, and the output is masks as binary matrices. The mask is used to extract the instance from the content image.

\subsection{Forward Stretching}
Topology is an important field of Mathematics, studying the properties of geometric objects. A key concept is that two objects are topologically identical (an equivalence relation) if they are homeomorphic. Formally speaking, if two topological spaces X and Y are \textbf{homeomorphic}, then there exists a bijective mapping $f:X\rightarrow Y$ such that both $f$ and $f^{-1}$ are continuous. The mapping $f$ is called the \textbf{homeomorphism} \cite{topology}. Intuitively, $f$ provides a one-to-one mapping between X and Y and $f^{-1}$ enables us to do it backward, from Y to X. As an example, a circular disk and a rectangular disk are homeomorphic because there exist $f$ and $f^{-1}$ that can transform a circular disk to a rectangular disk and vice versa.

Such an idea of deformation of objects motivates us to approach the problem of instance style transfer differently. Instead of transferring style onto the whole image and just extracting the area covered by the mask, we stretch the instance to rectangular shape, which can be represented as a tensor, and then transferring style onto the stretched instance. However, at the core of this process, there are two problems:

(a) There is an uncountable number of points on objects in two-dimensional space while images have only finite number of pixels.

(b) An explicit mapping $f$ needs to be defined.\\

  Mathematically, (a) can be rephrased as the following: given two points $x_0,x_1\in \mathbb{R}$ and the values of the functions at these points $g(x_0),g(x_1)$, what is the function value at some point $x\in [x_0,x_1]$? Linear interpolation provides a reasonable approximation by the polynomial of order one:
\begin{equation}
g(x)\approx P_{x_0,x_1}(x) := g(x_0)+\frac{g(x_1)-g(x_0)}{x_1-x_0}(x-x_0)\,,
\end{equation}
Although there is only finite number of pixels in an image, we can interpolate the value at any position as long as we have two pixels on its two sides.\\ 

\textbf{Forward Stretching} 
    An important insight in helping us develop Forward Stretching is Jordan–Schoenflies theorem, which states that the existence of homeophorism between an object in $\mathbb{R}^2$ with a simple closed curve as its boundary and a rectangular disk is guaranteed. This theorem provides theoretical backing on the shape of our output. Hence, we propose Forward Stretching with boolean mask $M\in \mathbb{R}^{H\times W}$ and content image $I_C\in \mathbb{R}^{H\times W\times C}$ as inputs and Stretched Instance $M^{++}$ in tensor form as output, where $H$, $W$, and $C$ are height, width, and number of channels of $I_C$. In essence, Forward Stretching maps pixels in each row in $I_C$ covered by $M$ to approximately evenly spaced slots and interpolates the values between these pixels. In addition, we recognize the fact that over-interpolation may negatively affect the quality of output, especially when instance as input is usually small. Therefore, in Forward Stretching, we introduce the concept of packing box, which is the smallest

\begin{algorithm}
\caption{Forward Stretching}
\label{alg1}
\textbf{inputs}:\multiline{%
Content Image: $I_C\in \mathbb{R}^{H\times W\times C}$\\
Mask: Boolean Matrix $M\in \mathbb{R}^{H\times W}$}
\begin{algorithmic}[1]
\Function{Forward}{M,$I_C$}
\State \multiline{%
Get coordinates of True values in M into 1d array $coordinates = \{(x_i,y_i)\}_{i\in \{1, 2,\ldots, N\}}$}
\State \multiline{%
Let $a = \min_i\{x_i\}$, $b = \max_i\{x_i\}$, $c = \min_i\{y_i\}$, and $d = \max_i\{y_i\}$}
\State \multiline{%
Crop $I_C$ into $I_C^{+}$ where each channel has vertices: $(a,c), (a, d), (b,c), (b,d)$}
\State \multiline{%
Crop M into packing box $M^{+}$ with vertices: $(a,c), (a, d), (b,c), (b,d)$}
\State \multiline{%
Let $P_h = d-c+1$ and $P_w = b-a+1$. $M^{+}\in \mathbb{R}^{P_h\times P_w}$ and $I_C^{+}\in \mathbb{R}^{P_h\times P_w\times C}$}
\State \multiline{%
Create Stretched Instance: $M^{++}\in \mathbb{R}^{P_h\times P_w\times C}$ filled with placeholder values}
\For {each channel c in $I_C^{+}$}
\For{each row r in $M^{+}$}
\State \multiline{%
Find corresponding Row j (\ie r = j) in c covered by r} 
\State \multiline{%
Let Row j have pixels $(x_1,y_j),\ldots,(x_{n_j},y_j)$ in increasing order of x coordinates and $n_j$ pixels covered by True values in r, where all x and y coordinates correspond to True values in row r} 
\State \multiline{%
 Place these pixels at $(a, y_j),(\floor*{a+{\frac{b-a}{n_j-1}}}, y_j),(\floor*{a+{\frac{2(b-a)}{n_j-1}}}, y_j),$\\$\ldots, (b, y_j)$, in corresponding Row k (\ie r = j = k) of $M^{++}$}
\State \multiline{%
Use Equation (1) to interpolate values between pixels in Row k of $M^{++}$ to replace placeholder values}
\EndFor
\EndFor
\EndFunction
\end{algorithmic}
\textbf{output:} Stretched Instance: $M^{++}\in \mathbb{R}^{P_h\times P_w \times C}$
\end{algorithm} 

\noindent rectangle that can cover the instance, to minimize this issue. Packing box designates the area in which the instance can be stretched in Forward Stretching. For details of Forward Stretching, please see Algorithm \ref{alg1}.

\subsection{Instance style transfer}
After Forward Stretching, the instance is in tensor form, and we apply the WCT Neural Network \cite{Universal} to transfer style to the stretched instance. WCT Neural Network in essence is an auto-encoder that inverts the feature maps of the content image after these feature maps are transformed by whitening and coloring. The encoder in our WCT Neural Network is VGG-19 \cite{Very}, and the decoders are symmetrical to the encoder and are trained with the loss function:
\begin{equation}
L = ||I_o - I_i||_2^{2}+\lambda ||\Phi(I_o) - \Phi(I_i)||_2^2\,,
\end{equation}
where $I_i$, $I_o$,  $\lambda$, and $\Phi$ are input image,  output image, the parameter to balance reconstruction loss and feature loss, and the VGG encoder that extracts features from ReluX\_1 (X = 1,2,3,4,5) layers, respectively \cite{Universal}. 

Whitening removes the style of the instance and coloring transfers the style from the style image to the whitened instance. Let's denote $f_c \in \mathbb{R}^{C \times M_c}$ and $f_s \in \mathbb{R}^{C \times M_s}$ as the vectorized feature maps of the instance and style image, respectively, extracted from the encoder, where $C$ is the number of channels of the feature maps and $M_c$ and $M_s$ are the products of height and width of the instance and style feature maps. We start with mean-centering $f_c$ and $f_s$ by their respective row vectors. The whitening transformation performs the following operation:
\begin{equation}
\hat{f_c} = E_cD_c^{-\frac{1}{2}}E_c^\top f_c\,,
\end{equation}
where, given $f_cf_c^\top = E_cD_cE_c^\top$, $E_c$ and $D_c$ are the orthogonal matrix of eigenvectors and diagonal matrix of eigenvalues of covariance matrix $f_cf_c^\top$, respectively \cite{Universal}.

The coloring transformation performs the following operation:
\begin{equation}
\hat{f}_{cs} = E_sD_s^{\frac{1}{2}}E_s^\top \hat{f_c}\,,
\end{equation}
where $E_s$ and $D_s$ are the orthogonal matrix of eigenvectors and diagonal matrix of eigenvalues of covariance matrix $f_sf_s^\top$, respectively \cite{Universal}.
  
  This model allows user to balance the blending between style and content with the following operation:
\begin{equation}
\hat{f}_{cs} = \alpha \hat{f}_{cs} + (1 - \alpha)f_c\,,
\end{equation}

$\hat{f}_{cs}$ is then fed to the decoder, and the stylized object in tensor form $O^{+}\in \mathbb{R}^{P_h\times P_w\times C}$ is returned.

  With five different decoders trained by Equation (2) ($\Phi$ extracts ReluX\_1 features where X = 1,2,3,4,5), as the original paper \cite{Universal} pointed out, we can build a multi-style pipeline where the output of the models using decoders trained with ReluX\_1 from higher levels will be used as input for models using decoders trained with ReluX\_1 from lower levels. In our method, we use all five decoders for multi-level stylization.
  
\subsection{Backward Stretching}
    After performing style transfer on the tensor achieved by Forward Stretching, we need to unstretch the object so that it can return to its original shape. This is accomplished by returning each original pixel with its new value (after style transfer) to its original position and increasing the size of the unstretched, stylized instance back to the size of the original content image $I_C$ by padding with zeros. For details of Backward Stretching, please see Algorithm 2.
     
\begin{algorithm}
\caption{Backward Stretching}
\label{alg2}
\textbf{inputs}:\multiline{%
Stylized Object: $O^{+}\in \mathbb{R}^{P_h\times P_w\times C}$}
\begin{algorithmic}[1]
\Function{Backward}{$O^{+}$}
\State \multiline{%
Replace every pixel that was obtained by interpolation in Forward Stretching with zero}
\For {each channel c in $O^{+}$}
\For{each row r in c}
\State \multiline{%
Move non-interpolated pixels from coordinates obtained from Forward Stretching, $(a,y_r),(\floor*{a+{\frac{b-a}{n_r-1}}},y_r),(\floor*{a+{\frac{2(b-a)}{n_r-1}}},y_r),$\\$\ldots, (b,y_r)$, back to $(x_1,y_r),\ldots,(x_{n_r},y_r)$, the original positions}
 
\EndFor
\EndFor
\State \multiline{%
Pad the tensor with zeros so that it has the same shape as $I_C$}

\EndFunction
\end{algorithmic}
\textbf{output:} Stylized Instance: $O^{++}\in \mathbb{R}^{H\times W \times C}$
\end{algorithm}

After Backward Stretching, let's denote $I_C^{++}\in \mathbb{R}^{H\times W \times C}$ as the image where the pixels covered by the mask are all replaced by zeros. The final output $I$ is obtained by:
\begin{equation}
I = I_C^{++} + O^{++}\,,
\end{equation}

\noindent where $I\in \mathbb{R}^{H\times W \times C}$

\section{Experimental results}
Due to the small number of works and thus available models in this field, we only showcase the results of our method in Figure.3 for qualitative evaluation. We selected images with instances from different categories in different sizes to demonstrate the effectiveness of our method.

\begin{figure*} 
\centering 
\includegraphics[width=16cm, height=22cm]{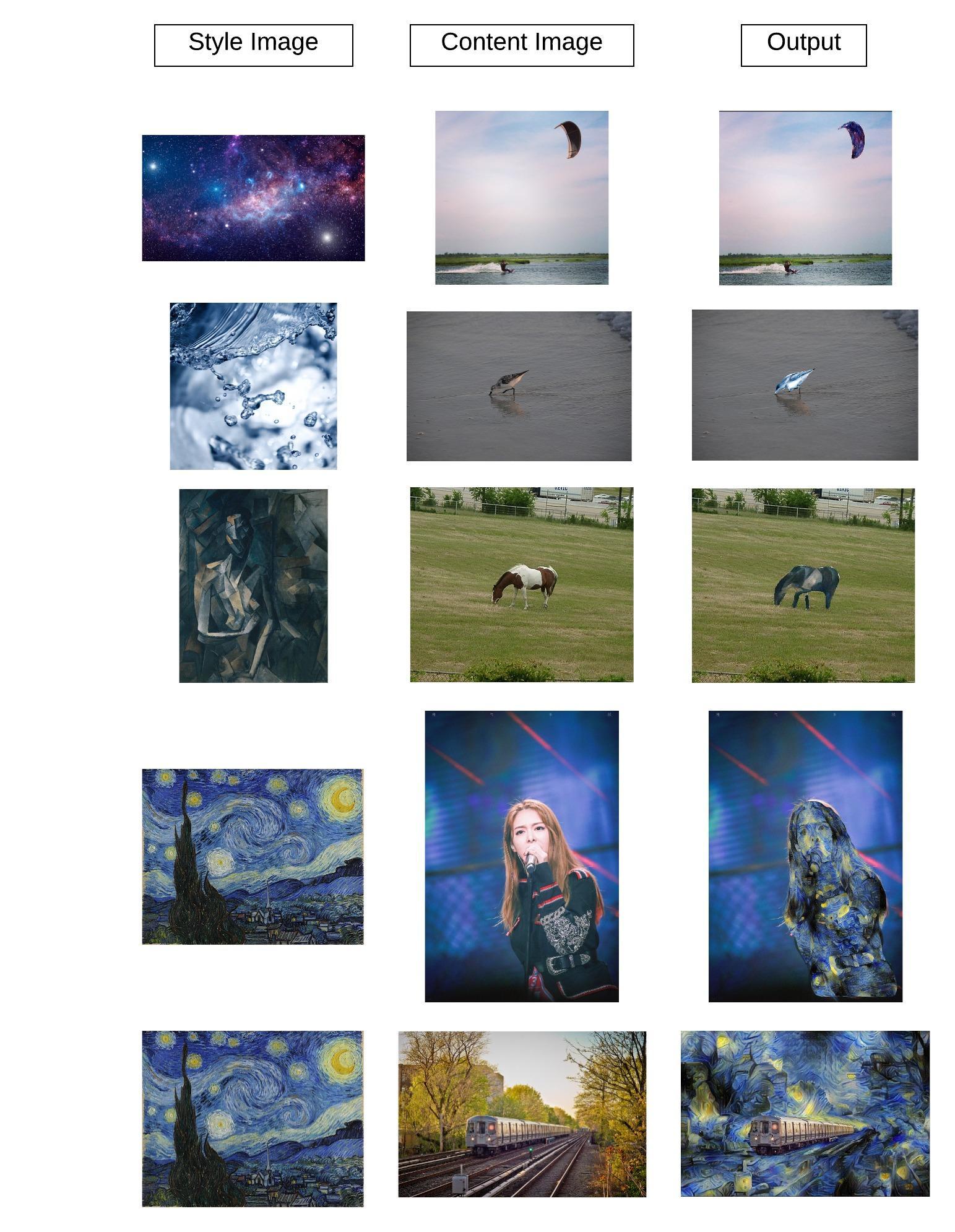} 
\caption{\textsf{Outputs from our method. We select content images with instances in small sizes (\eg bird and parachute) and large size (\eg person) and transfer style to instances as well as background to demonstrate possible applications of instance style transfer and our method.}} 
\label{fig:ds} 
\end{figure*} 

\section{Discussion}

  Our method comprises three main parts: instance segmentation, Forward Stretching, and image style transfer. The improvements in each of these sub-tasks will enhance the quality of outputs. Firstly, the quality of the mask (\ie if it is covering the entire instance) determines the area into which the style transfers and thus the extent of visual coherence of the instance after instance style transfer. We also hypothesize that a three-dimensional mask, which also captures depth information and represents the instance more realistically, should help produce better outputs for instance style transfer. Secondly, the algorithm used to transform instance into a tensor is critical. In this work, we only explored one-dimensional stretching of instance with linear interpolation. Future works can study other interpolation methods, two-dimensional stretching of instance, or other transformations to handle instances in arbitrary shapes. Lastly, instance style transfer, as the name implies, works with instance, which is on a smaller scale than an entire image. The capability of image style transfer models to preserve and transfer style to tiny details of the instance is much more pivotal in the success of instance style transfer than in image style transfer. Therefore, image style transfer models fine-tuned to small details of content images are uniquely positioned in instance style transfer.
  
  Drawing from our experience in this project, we discovered that the visual appealingness of the content image, after the stylized instance is integrated back to it, is determined by the quality of the stylized instance as well as the color and texture of the area that is not stylized. If the color and texture of the stylized instance are substantially distinct from those of the area untouched by our method, the stylized instance will look incongruous. Thus, further works can be done in understanding and balancing the difference between the colors and textures of targeted instance and non-targeted area. Such works would increase machine intelligence to provide context and even help make recommendations for instance style transfer.
  
  Finally, instance style transfer rests on the assumption that the instance of interest can be segmented and extracted from the content image, and we believe this is one of the most crucial factors in determining how widely in the future instance style transfer using machine learning will be used in industries. That may result in a trade-off between choosing a model that can segment a wide variety of instances and a model that can segment instances precisely. In any case, our method provides flexibility for users to replace Mask R-CNN with another instance segmentation model and WCT Neural Network with another image style transfer model.
\section{Conclusion}
  In this paper, we proposed a method for instance style transfer with insights from instance segmentation, topology, and image style transfer, which allows us to transfer arbitrary style to an instance in an arbitrary shape. We introduced Forward Stretching to transform instance into a tensor with theoretical backing from homeomorphic spaces in topology. Finally, we presented our results to demonstrate that our method can perform well in real-world images with instances in various sizes and categories.

% Mask R-CNN [4], WCT Neural Network [7] 
% https://www.apple.com/newsroom/2016/12/apple-unveils-best-of-2016-across-apps-music-movies-and-more/ 

% Mask r cnn 

% https://arxiv.org/pdf/1701.02357.pdf 

% http://cs231n.stanford.edu/reports/2017/pdfs/416.pdf 

% https://arxiv.org/pdf/1705.08086.pdf 

% https://arxiv.org/pdf/1508.06576.pdf 

% https://arxiv.org/pdf/1511.04517.pdf 

% The followings are found in Path Aggregation Network for Instance Segmentation 

%  https://arxiv.org/pdf/1512.04412.pdf 

%  https://arxiv.org/pdf/1803.01534.pdf 

%  https://arxiv.org/pdf/1611.08303.pdf 

%  http://www.cs.toronto.edu/~fidler/papers/sgn_iccv17.pdf 

% https://arxiv.org/pdf/1604.05096.pdf 

% https://arxiv.org/pdf/1603.08155.pdf 

% https://arxiv.org/pdf/1603.03417.pdf 

% https://arxiv.org/pdf/1703.09210.pdf 

% https://arxiv.org/pdf/1610.07629.pdf 

% https://arxiv.org/pdf/1908.11525.pdf 

% https://arxiv.org/pdf/1612.03144.pdf 

% https://arxiv.org/pdf/1409.1556.pdf 

% https://arxiv.org/pdf/1512.03385.pdf 

\printbibliography

@article{2017, title={Arbitrary Style Transfer in Real-Time with Adaptive Instance Normalization}, author={Xun Huang and Serge J. Belongie}, journal={2017 IEEE International Conference on Computer Vision (ICCV)}, year={2017}, pages={1510-1519} }

@manual{Senn,
      title  = "Apple unveils Best of 2016 across apps, music, movies and more",
      url    = "https://www.apple.com/newsroom/2016/12/apple-unveils-best-of-2016-across-apps-music-movies-and-more/",
      year   = "December 6, 2016 "
    }

@misc{classbased,
    title={Class-Based Styling: Real-time Localized Style Transfer with Semantic Segmentation},
    author={Lironne Kurzman and David Vazquez and Issam Laradji},
    year={2019},
    eprint={1908.11525},
    archivePrefix={arXiv},
    primaryClass={cs.CV}
}

@book{topology,
  title={Topology},
  author={Munkres, J.R.},
  isbn={9780131816299},
  lccn={99052942},
  series={Featured Titles for Topology Series},
  url={https://books.google.com/books?id=XjoZAQAAIAAJ},
  year={2000},
  publisher={Prentice Hall, Incorporated}
}

@INPROCEEDINGS{son,
author={C. {Castillo} and S. {De} and X. {Han} and B. {Singh} and A. K. {Yadav} and T. {Goldstein}},
booktitle={2017 IEEE International Conference on Acoustics, Speech and Signal Processing (ICASSP)},
title={Son of Zorn's lemma: Targeted style transfer using instance-aware semantic segmentation},
year={2017},
volume={},
number={},
pages={1348-1352},
doi={10.1109/ICASSP.2017.7952376},
ISSN={},
month={March},}

@incollection{Universal,
title = {Universal Style Transfer via Feature Transforms},
author = {Li, Yijun and Fang, Chen and Yang, Jimei and Wang, Zhaowen and Lu, Xin and Yang, Ming-Hsuan},
booktitle = {Advances in Neural Information Processing Systems 30},
editor = {I. Guyon and U. V. Luxburg and S. Bengio and H. Wallach and R. Fergus and S. Vishwanathan and R. Garnett},
pages = {386--396},
year = {2017},
publisher = {Curran Associates, Inc.},
url = {http://papers.nips.cc/paper/6642-universal-style-transfer-via-feature-transforms.pdf}
}

@INPROCEEDINGS{Image,
author={L. A. {Gatys} and A. S. {Ecker} and M. {Bethge}},
booktitle={2016 IEEE Conference on Computer Vision and Pattern Recognition (CVPR)},
title={Image Style Transfer Using Convolutional Neural Networks},
year={2016},
volume={},
number={},
pages={2414-2423},
doi={10.1109/CVPR.2016.265},
ISSN={},
month={June},}

@article{Reversible,
  title={Reversible Recursive Instance-Level Object Segmentation},
  author={Xiaodan Liang and Yunchao Wei and Xiaohui Shen and Zequn Jie and Jiashi Feng and Liang Lin and Shuicheng Yan},
  journal={2016 IEEE Conference on Computer Vision and Pattern Recognition (CVPR)},
  year={2015},
  pages={633-641}
}

@inproceedings{Instance,
author = {Dai, Jifeng and He, Kaiming and Sun, Jian},
year = {2016},
month = {06},
pages = {3150-3158},
title = {Instance-Aware Semantic Segmentation via Multi-task Network Cascades},
doi = {10.1109/CVPR.2016.343}
}

@article{Path,
  title={Path Aggregation Network for Instance Segmentation},
  author={Shu Liu and Lu Qi and Haifang Qin and Jianping Shi and Jiaya Jia},
  journal={2018 IEEE/CVF Conference on Computer Vision and Pattern Recognition},
  year={2018},
  pages={8759-8768}
}

@article{Deep,
  title={Deep Watershed Transform for Instance Segmentation},
  author={},
  journal={2017 IEEE Conference on Computer Vision and Pattern Recognition (CVPR)},
  year={2016},
  pages={2858-2866}
}

@INPROCEEDINGS{SGN,
author={S. {Liu} and J. {Jia} and S. {Fidler} and R. {Urtasun}},
booktitle={2017 IEEE International Conference on Computer Vision (ICCV)},
title={SGN: Sequential Grouping Networks for Instance Segmentation},
year={2017},
volume={},
number={},
pages={3516-3524},
doi={10.1109/ICCV.2017.378},
ISSN={},
month={Oct},}

@inproceedings{perceptual,
  title={Perceptual losses for real-time style transfer and super-resolution},
  author={Johnson, Justin and Alahi, Alexandre and Fei-Fei, Li},
  booktitle={European Conference on Computer Vision},
  year={2016}
}

@inproceedings{StyleBank,
author = {Chen, Dongdong and Yuan, Lu and Liao, Jing and Yu, Nenghai and Hua, Gang},
year = {2017},
month = {07},
pages = {2770-2779},
title = {StyleBank: An Explicit Representation for Neural Image Style Transfer},
doi = {10.1109/CVPR.2017.296}
}

@inproceedings{ALearned,
author = {Dumoulin, Vincent and Shlens, Jonathon and Kudlur, Manjunath},
year = {2017},
month = {04},
title = {A Learned Representation For Artistic Style},
series = {ICLR'17}
}

@inproceedings{Texture,
 author = {Ulyanov, Dmitry and Lebedev, Vadim and Vedaldi, Andrea and Lempitsky, Victor},
 title = {Texture Networks: Feed-forward Synthesis of Textures and Stylized Images},
 booktitle = {Proceedings of the 33rd International Conference on International Conference on Machine Learning - Volume 48},
 series = {ICML'16},
 year = {2016},
 location = {New York, NY, USA},
 pages = {1349--1357},
 numpages = {9},
 url = {http://dl.acm.org/citation.cfm?id=3045390.3045533},
 acmid = {3045533},
 publisher = {JMLR.org},
}

@article{Very,
  title={Very Deep Convolutional Networks for Large-Scale Image Recognition},
  author={Karen Simonyan and Andrew Zisserman},
  journal={CoRR},
  year={2014},
  volume={abs/1409.1556}
}

@inproceedings{DeepResidual,
author = {He, Kaiming and Zhang, Xiangyu and Ren, Shaoqing and Sun, Jian},
year = {2016},
month = {06},
pages = {770-778},
title = {Deep Residual Learning for Image Recognition},
doi = {10.1109/CVPR.2016.90}
}

@INPROCEEDINGS{July,
author={T. {Lin} and P. {Dollár} and R. {Girshick} and K. {He} and B. {Hariharan} and S. {Belongie}},
booktitle={2017 IEEE Conference on Computer Vision and Pattern Recognition (CVPR)},
title={Feature Pyramid Networks for Object Detection},
year={2017},
volume={},
number={},
pages={936-944},
doi={10.1109/CVPR.2017.106},
ISSN={},
month={July},}

@INPROCEEDINGS{Mask,
author={K. {He} and G. {Gkioxari} and P. {Dollár} and R. {Girshick}},
booktitle={2017 IEEE International Conference on Computer Vision (ICCV)},
title={Mask R-CNN},
year={2017},
volume={},
number={},
pages={2980-2988},
doi={10.1109/ICCV.2017.322},
ISSN={},
month={Oct},}
\end{document}